%% file: main.tex
\documentclass{article}

    \PassOptionsToPackage{numbers, compress}{natbib}

\usepackage[preprint]{neurips_2024}

\usepackage[utf8]{inputenc} %
\usepackage[T1]{fontenc}    %
\usepackage{hyperref}       %
\usepackage{url}            %
\usepackage{booktabs}       %
\usepackage{amsfonts}       %
\usepackage{nicefrac}       %
\usepackage{microtype}      %
\usepackage{xcolor}         %

\title{Weakly Supervised Point Cloud Segmentation via Conservative \ky{Propagation} of Scene-level Labels}
\input{preamble}

\author{%
    Shaobo Xia$^{1}$\thanks{Equal contribution.} \And
    Jun Yue$^{2}$\footnotemark[1] \And
    Kacper Kania$^{3}$ \And
    Leyuan Fang$^{4}$ \And
    Andrea Tagliasacchi$^{5, 6}$ \And Kwang Moo Yi$^{7}$ $\qquad$ Weiwei Sun$^{7, 8}$\thanks{Work performed while at the University of British Columbia.} 
    \\ [.2in]
    $^1$Changsha University of Science and Technology \hspace{1pt}
    $^2$Central South University \hspace{1pt} \\ 
    $^3$Warsaw University of Technology \hspace{1pt} 
    $^4$Hunan University \hspace{1pt}
    $^5$Simon Fraser University \hspace{1pt} \\
    $^6$University of Toronto \hspace{1pt}
    $^7$University of British Columbia \hspace{1pt} 
    $^8$Amazon \hspace{1pt} 
    \\[.2in]
}

\begin{document}
\maketitle
\input{sec/0_abstract}

\input{sec/1_intro}
\input{sec/2_related}

\input{sec/3_method}

\input{sec/4_experiments}

\input{sec/5_conclusions.tex}

\input{sec/6_acks}
{
  \bibliographystyle{ieeenat_fullname}
  \bibliography{main}
}

\newpage
\appendix
\input{sec/X_suppl}

\end{document}

%% file: preamble.tex
\usepackage{color}
\usepackage{booktabs}
\usepackage{multirow}
\usepackage{graphicx}
\usepackage{overpic}
\usepackage[utf8]{inputenc}\DeclareUnicodeCharacter{2212}{-} 

\makeatletter
\DeclareRobustCommand\onedot{\futurelet\@let@token\@onedot}
\def\@onedot{\ifx\@let@token.\else.\null\fi\xspace}

\def\eg{\emph{e.g}\onedot} 
\def\ie{\emph{i.e}\onedot}

\makeatother

\AtEndPreamble{
    \usepackage[capitalize]{cleveref}
    \crefname{section}{Sec.}{Secs.}
    \Crefname{section}{Section}{Sections}
    \Crefname{table}{Table}{Tables}
    \crefname{table}{Tab.}{Tabs.}
    \Crefname{equation}{Equation.}{Equations.}
    \crefname{equation}{Eq.}{Eqs.}
}

\definecolor{turquoise}{cmyk}{0.65,0,0.1,0.3}
\definecolor{purple}{rgb}{0.65,0,0.65}
\definecolor{dark_green}{rgb}{0, 0.5, 0}
\definecolor{orange}{rgb}{0.8, 0.6, 0.2}
\definecolor{red}{rgb}{0.8, 0.2, 0.2}
\definecolor{darkred}{rgb}{0.6, 0.1, 0.05}
\definecolor{blueish}{rgb}{0.0, 0.3, .6}
\definecolor{light_gray}{rgb}{0.7, 0.7, .7}
\definecolor{pink}{rgb}{0.9, 0, 0.6}
\definecolor{greyblue}{rgb}{0.25, 0.25, 1}
\definecolor{teal}{rgb}{0.0, 0.4, 0.4}

\usepackage{pgfplots}
\usepackage{tikz}
\usepackage{mathtools}
\usepackage{wrapfig}
\usepackage{caption}

\usetikzlibrary{matrix, positioning,arrows,arrows.meta,calc,decorations.pathreplacing,decorations.text,spy}
\pgfplotsset{compat=1.18}

\tikzset{
  img/.style={
    inner sep=0pt,     %
    outer sep=0pt,     %
    rectangle,
    align=center} %
}

\newcommand{\ky}[1]{{\color{dark_green}#1}}

\newcommand{\ws}[1]{{\color{pink}#1}}

\renewcommand{\ky}[1]{#1}
\renewcommand{\ws}[1]{#1}

\usepackage{soul}

\usepackage{dsfont}

\newcommand{\Eq}[1]{Eq.~\ref{eq:#1}}

\usepackage{blindtext}

\usepackage{enumitem}
\setlist[itemize]{noitemsep,leftmargin=*,topsep=.05in}
\setlist[enumerate]{noitemsep,leftmargin=*,topsep=.05in}

\def \customparskip {0.5em}
\renewcommand{\paragraph}[1]{\vspace{\customparskip}\noindent\textbf{#1}.}

\newcommand{\linear}{\mathcal{L}}

\DeclareMathOperator*{\argmin}{arg\,min}
\DeclareMathOperator*{\argmax}{arg\,max}
\newcommand{\loss}[1]{\ell_\text{#1}}

\newcommand{\param}{\boldsymbol{\theta}}
\newcommand{\paramlinear}{\boldsymbol{\omega}}

\newcommand{\real}{\mathbb{R}}

\newcommand{\features}{\mathbf{f}}

\newcommand{\pars}{\param}

\newcommand{\prim}{\ensuremath{\text{prim}}}
\newcommand{\y}{\ensuremath{\mathbf{y}}}

\newcommand{\ypoint}{\ensuremath{\mathbf{y}^{\text{point}}}}

\newcommand{\entropy}{\ensuremath{\mathcal{H}}}

\newcommand{\affinity}{\ensuremath{\mathbf{A}}}

\newcommand{\affinityprim}{\ensuremath{\affinity^{\text{prim}}}}

\newcommand{\featuresprim}{\ensuremath{\features^\prim}}
\newcommand{\scores}{\ensuremath{\mathbf{s}}}
\newcommand{\ypoints}{\ensuremath{\mathbf{y}^\text{point}}}
\newcommand{\supplementary}{\texttt{Supplementary}}

\newcommand{\avgfeaturesprim}{\ensuremath{\bar{\features}^\prim}}

%% file: sec/0_abstract.tex
\begin{abstract}
We propose a weakly supervised semantic segmentation method for point clouds that predicts ``per-point'' labels from just ``whole-scene'' annotations.
The key challenge here is the discrepancy between the target of dense per-point semantic prediction and training losses derived from only scene-level labels. 
\ky{%
To address this, in addition to the typical weakly-supervised setup that supervises \emph{all points} with the scene label, we propose to \emph{conservatively propagate} the scene-level labels to points selectively.
Specifically, we over-segment point cloud features via unsupervised clustering in the entire dataset and form primitives.
We then associate scene-level labels with primitives through bipartite matching.
Then, we allow labels to pass through this primitive--label relationship, while further encouraging features to form narrow clusters around the primitives.
Importantly, through bipartite matching, this additional pathway through which labels flow, only propagates scene labels to the most relevant points, reducing the potential negative impact caused by the global approach that existing methods take.
}%
We evaluate our method on ScanNet and S3DIS datasets, outperforming the state of the art by a large margin.
\end{abstract}

%% file: sec/1_intro.tex
\section{Introduction}
Semantic segmentation of point clouds is a core problem in 3D Computer Vision~\cite{peng2023openscene, sun2023neuralbf,guo2020deep}.
It is the foundation of many applications, including scene understanding~\cite{peng2023openscene}, semantic reconstruction~\cite{nie2021rfd}, and urban mapping~\cite{hu2022sensaturban}.
Many state-of-the-art methods~\cite{Mix3D, OctFormer, choy20194d} rely on dense, \textit{per-point} supervision which requires intense manual labour.
For example, annotating ScanNet~\cite{dai2017scannet} took 500 workers an average of 22.3 minutes per scene, and there are 1513 scenes in the dataset (${\approx}1$ month of 24/7 annotation time).
To reduce this effort, weakly-supervised learning~ \cite{XuTowards2020, hu2021sqn, unal2022scribble, tao2022seggroup, LESS2022,Wei_2020_CVPR} can be used, which assumes access to much fewer annotations than the fully supervised counterparts.

Among weakly-supervised methods, in this work, we focus on \textit{scene-level} labels, where the training signal is the existence of a given class within the scene.
Scene labels are 
cheaper to obtain than dense ones, as each category takes only~${\approx}1$ second to annotate~\cite{papadopoulos2014training}.
This means the annotation cost of a single scene in ScanNet~\cite{dai2017scannet} is reduced~$20\times$, from ${\approx}20$ minutes, to roughly~${\approx}1$ minute~\cite{yang2022mil}.
Further, scene-level labels do not rely on specialized 3D annotation software or extensive annotator training.

While recent state-of-the-art methods for scene-level supervised point cloud segmentation differ in architecture~\cite{yang2022mil, mit_2023, ren20213d, Wei_2020_CVPR}, to our best knowledge most of them rely on Class Activation Mapping (CAM)~\cite{zhou2016learning}.
CAM trains a global classification network with the scene-level labels and \ky{with an architectural bottleneck -- global pooling -- that allows the resultant network to perform point-level segmentation with small modifications during inference.}
\ky{Specifically, as shown in \cref{fig:grad}, CAM utilizes global average pooling to convert point-wise class scores into a global score distribution for classification during training, where the pooling can be dropped during inference, thus categorizing each point directly.}
However, under this formula of global classification, all points within a scene share the same ground-truth scene-level label -- 
\ky{all points are affected by the global label, which can be detrimental.
For example, the label for the chair will also affect the training of features extracted for the table.
While in theory this should be mitigated by the high dimensionality of the point cloud features, in practice this is not the case and leads to suboptimal performances that are much worse than those with full supervision.}

\input{figs/grad}

In this work, in addition to the typical weak-supervision pathway, we thus propose an alternative path for \emph{confident} labels to flow through, to only those points that are relevant.
Specifically, we first \textit{over-segment} the point cloud feature space via K-means~\cite{lloyd1982least} into \textit{feature centroids}~\cite{zhang2023growsp}, which we call \textit{primitives} across the entire dataset.
Then, for each scene, we associate a subset of these primitives with the scene labels through bipartite matching~\cite{kuhn1955hungarian}.
Note that with bipartite matching, we create a restricted relationship between labels and points that is much more conservative -- each class matches only one primitive with the highest confidence and vice versa.
We then let labels propagate through this pathway to associated 
\input{figs/teaser/teaser_graph_only}points in the point cloud.
Finally, we further encourage features to form narrow clusters around primitives to facilitate this process.
Note that this label propagation pathway that the primitive--label pairs provide now only covers a few parts of the scene, but acts as \emph{anchors} for the entire weakly-supervised training to rely on.
Moreover, as shown in \cref{fig:grad}, for this pathway, labels no longer affect all points but only those that are matched.

With our method, we achieve new state-of-the-art results on both ScanNet~\cite{dai2017scannet} and S3DIS~\cite{armeni20163d}, significantly surpassing the previous state of the art respectively by 8.9\% and 4.8\% in terms of absolute mean Intersection over Union (mIoU), which, again, respectively \emph{corresponds to a leap in performance of 29\% and 18\% relative gain}; see~\cref{fig:teaser_graph}.
In fact, for the first time, our method outperforms certain fully supervised methods (see \cref{tab:scannet}), highlighting the real potential of weakly supervised learning in point clouds.

In summary, in this paper:
\begin{itemize}[leftmargin=*]
\item \ky{we propose a novel method to weakly-supervised semantic segmentation of point clouds that augments an additional \emph{conservative} pathway for labels to propagate;}
\item \ky{to construct this pathway, we propose to use bipartite matching, hence creating label--cluster pairs using unsupervised clustering;}
\item \ky{we then propose to learn features such that the average features of each cluster are correctly classified, thus locally propagating the labels only to each corresponding cluster;}
\item \ky{our approach significantly outperforms the state of the art in weakly supervised point cloud segmentation.}    
\end{itemize}

%% file: figs/grad.tex
\begin{figure}[t]
  \includegraphics[width=\linewidth]{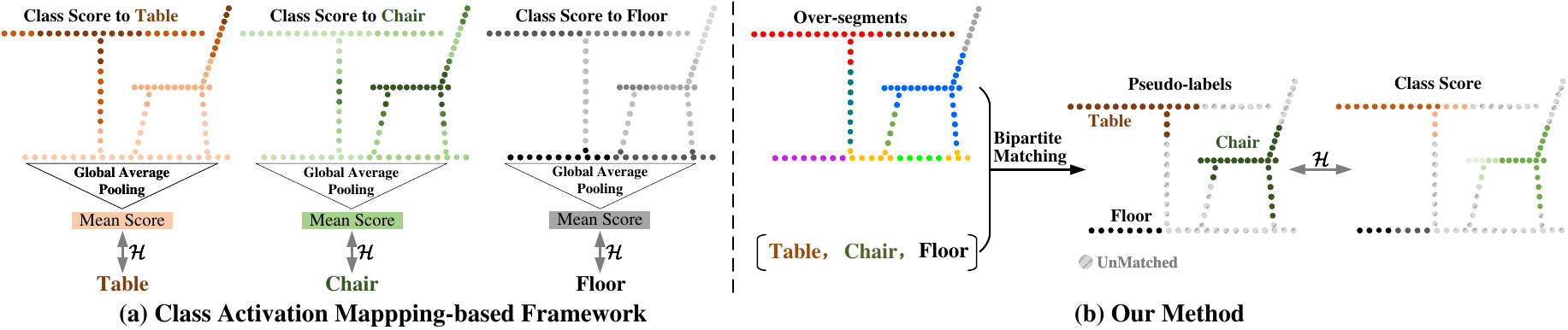}
  \caption{
    \textbf{Pitfall of Class Activation Mapping (CAM)~\cite{zhou2016learning}-based methods} -- 
    \ky{%
    (a) Methods based on class activation mapping rely on global average pooling, which then propagates the label to points `globally', thus all points are affected by all labels.
    (b) Our method, on the other hand, first over-segments the point cloud and then propagates labels through bipartite matching with these clusters (primitives), thus only propagating labels to those points that are most relevant, ultimately leading to a better point cloud segmentation performance.
    }%
  }
  \label{fig:grad}
  \vspace{-1.8em}
\end{figure}

%% file: figs/teaser/teaser_graph_only.tex
\begin{wrapfigure}{r}{0.3\linewidth}
    \vspace{-1.5em}
    \begin{center}
    \includegraphics[width=\linewidth]{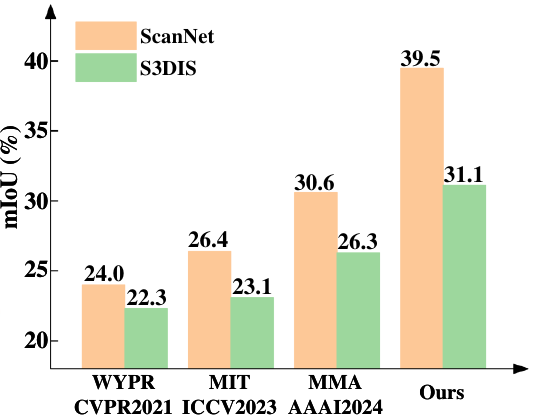}
    \end{center}
    \vspace{-1em}
    \caption{
    {\bf Results preview} -- 
    Performance of various baselines and our method on ScanNet~\cite{dai2017scannet} and {S3DIS}~\cite{armeni20163d}.
    Ours significantly outperforms the state of the art.
    }
    \vspace{-1.5em}
    \label{fig:teaser_graph}
\end{wrapfigure}

%% file: sec/2_related.tex
\section{Related works}

\vspace{-\customparskip}
\paragraph{Weakly supervised point cloud segmentation}
This task aims to achieve point cloud semantic segmentation with a significantly reduced labeling cost.
Most commonly, a few scenes in a dataset have full annotations, and most of the other scenes have no labels~\cite{jiang2021guided, zhao2020sess, gao2023dqs3d,kong2023lasermix}.
Exploiting unannotated data's full potential is key to obtaining high-quality results~\cite{jiang2021guided, ChengSSPC2021, xu2023hierarchical}.
Other works assume that only 0.1\%-10\% points need to be annotated~\cite{XuTowards2020, liu2021oneclick, hou2021exploring, li2022hybridcr, WS3D2022, zhang2021perturbed, zhang2021weakly, xu2022back, MultiPrototype}.
Alternatively, one can annotate bounding boxes~\cite{chibane2022box2mask}, scribbles~\cite{unal2022scribble}, pre-processed segments~\cite{tao2022seggroup} and clusters~\cite{LESS2022}.
As there are direct connections between points and semantic labels, they mainly focus on propagating reliable labels to unknown regions~\cite{hu2021sqn, liu2023clickseg, liu2023cpcm}.
For instance,~\citet{hu2021sqn} explores how labels propagate at different scales, enabling unlabeled points to utilize point-level annotations.
Nevertheless, these approaches still need accurate and costly point-level labels, while weakly supervised semantic segmentation with \textit{scene-level} annotations only requires listing the category names in a scene~\cite{Wei_2020_CVPR}.
Scene-level annotations provide more opportunities for low-cost labeling, including leveraging cross-modal data (\eg text, speech, images), and our research thus focuses on this direction.

\paragraph{CAM in point clouds}
The critical challenge of scene-level weakly supervised learning lies in the absence of a direct connection between semantic tags and points. Class Activation Mapping~(CAM), aided by global average pooling~(GAP), is an essential tool to bridge per-point and scene-level tags~\cite{zhou2016learning}.
\citet{Wei_2020_CVPR} introduce~CAM into weakly supervised point cloud segmentation.
\citet{ren20213d} use a bottom-up point merging algorithm to merge points within each scene into pseudo-instances, thereby improving the association between semantic objects and scene tags under CAM. \citet{yang2022mil} develop a CAM-based weakly supervised semantic segmentation framework based on Transformers~\cite{vaswani2017attention}.
They improve the mapping relationship by considering the proportions of different objects in a scene through a weighted GAP.
Additionally, \citet{mit_2023} show that super-voxels used as tokens in Transformers can further improve the semantic accuracy. MMA~\cite{li2024multi} enhances the CAM-based framework by optimizing the feature consistency of different modalities, such as color and geometric shapes. Like~\citet{ren20213d}, it also employs pseudo-labels for the majority of points to facilitate self-training. \ws{However, these CAM-based approaches are known for supervising all points with the same multi-hot scene-level label, even though a point should be assigned to a single class only. Our work addresses this core limitation.}

\paragraph{Unsupervised learning}
Unsupervised learning can learn primitives (or parts) that contain semantic patterns without any annotations~\cite{cho2021picie, jin2022tusk}.
In point clouds, object-level segmentation via unsupervised learning has been studied extensively~\cite{sauder2019self, sun2020canonical}. Several methods~\cite{xie2020pointcontrast, hou2021exploring} also over-segment point clouds into clusters based on the learned discriminative features, however, those clusters are error-prone as shown by~\citet{zhang2023growsp}.
\citet{zhang2023growsp} proposes a practical unsupervised method based on K-means, and achieves state-of-art unsupervised semantic segmentation for point clouds.
As the key to the superior performance, their method explores cross-scene prior by performing K-means in the entire training dataset, \ie, a strategy which is extensively used in the image domain~\cite{cho2021picie, zhou2022regional,jo2023mars, crossimage2023}.
While we leverage these advances to obtain primitives, we experimentally show that combining these with scene-level supervision is not trivial, and our method addresses this very problem.

\paragraph{Bipartite matching}
Bipartite matching pairs elements from two distinct sets~\cite{kuhn1955hungarian}.
It finds the optimal subset of edges between two disjoint sets of vertices in a bipartite graph that minimizes the assignment cost.
It has been used with success in recent methods for object detection~\cite{carion2020end}, bounding box annotation~\cite{stewart2016end}, and unsupervised object discovery~\cite{locatello2020object}.
The optimal assignment between two sets can also be regarded as an optimal transport problem~\cite{sarlin2020superglue, caron2020unsupervised, jin2022tusk}.
We use bipartite matching to align learned primitives with scene-level tags.
With such a design, we achieve a new state-of-the-art in scene-level weakly supervised semantic segmentation.

%% file: sec/3_method.tex
\newcommand{\dataset}{\mathcal{P}}
\newcommand{\cloud}{\mathbf{P}}
\newcommand{\segmentation}{\mathbf{C}}

\input{figs/framework}

\section{Method}
In \Cref{fig:framework}, we provide an overview of our weakly-supervised approach to point cloud segmentation.
During training, we use only \textit{scene-level} labels -- those that indicate whether objects of a given class exist within the given scene.
Formally, given a dataset $\dataset = \{ \cloud_n \}$ of~$N$ point clouds, where $\cloud_n \in \real^{M_n\times 3}$, and the scene-level multi-hot labels $\mathcal{Y}=\{ \y_n \}$, where $\y_n \in \{0, 1\}^{C}$ indicates whether the \mbox{$n$-th} scene contains points belonging to one~(or more) of the $C$ categories, we aim to predict dense semantic segmentation~$\mathcal{C} = \{ \segmentation_n \}$, where $\segmentation_n \in \real^{M_n \times C}$.
\ky{To allow our method to be applicable to point clouds with an arbitrary number of points, we do not make any}
assumptions about the cardinality of the point cloud $M_n$.
\ky{Further, we drop the index $n$ for simplicity in notation,} without loss of generality.
\ky{We then aim} to find the optimal set of parameters~$\pars$\footnote{With~$\pars$, we generally denote \textit{all} the trainable parameters of our method.} that minimize losses:
\begin{align}
  \argmin_{\pars} \:\:
  \underbrace{\loss{cam}(\pars)}_\text{\Cref{sec:cam_preliminary}}
  + \underbrace{\loss{us}(\pars)}_\text{\Cref{sec:clustering}}
  + \underbrace{\loss{match}(\pars)}_{\text{\Cref{sec:assignment}}}
  .
  \label{eq:our-loss}
\end{align}%
Our core contributions are the latter two losses, \ky{that provide an alternative pathway for labels to influence training that is conservative and allows a significant performance gain.}

\subsection{Brief Review of Class Activation Maps (CAM)}
\label{sec:cam_preliminary}
\ky{%
For completeness, we first briefly review Class Activation Maps (CAM)~\cite{zhou2016learning, yang2022mil,mit_2023}
The core idea of CAM is to learn a point segmentation network through scene-level labels, with the help of global pooling.
}%
Specifically, given $H$-dimensional point features~$\features \in \real^{M\times H}$ computed from an off-the-shelf backbone network~\cite{graham20183d}, a \textit{linear} layer $\linear$ with trainable weights $\paramlinear \in \real^{H \times C}$ is trained to predict point-wise class scores~$\scores \in \real^{M\times C}$ as~$\scores = \linear(\features; \paramlinear)$.
\ky{%
Then, to supervise this training with per-scene class labels~$\y \in \{0, 1\}^C$, global average pooling is used.
We thus write
}%
\begin{align}
\loss{cam} = \mathcal{H}_C\biggl(
\underbrace{
  \frac{1}{M}\sum_{m=1}^M \scores_m
}_{\text{average pooling}} , \:\:
\y
\biggr)
,
\label{eq:lcam}
\end{align}
where $\y$ is \textit{multi-hot}, \ky{denoting the existence of multiple classes,} and $\entropy_C$ is the sum of binary entropies over $C$ classes:
\begin{align}
\mathcal{H}_C(\scores, \y)  = \sum_{c=1}^{C} \mathcal{H}(\bar{\scores}_c, \: y_c), 
\quad 
\bar{\scores}_c = \frac{1}{M}\sum^M_{m=1}\scores_{m, c}
\label{eq:mil}
.
\end{align}
\ky{%
This is different 
}%
from the mean entropy objective employed by \textit{dense} (fully-supervised) methods:
\begin{align}
  \loss{dense} =
    \frac{1}{M}\sum_{m=1}^M
    \mathcal{H}(\scores, \: \ypoints_m), \quad \forall_{m \in M}\| \ypoints_m  \|{=}1 
    ,
  \label{eq:ldense}
\end{align}
where $\ypoints_m$ is a one-hot label vector, and $\mathcal{H}$ is the simple cross entropy function.
\ky{%
The main difference between \cref{eq:ldense} and \cref{eq:lcam} is the supervision -- one is per scene, the other is per point. 
In \cref{eq:lcam} the fact that $\ypoints_m$ does not exist is bypassed by the global average pooling inside the entropy calculation.
}%
This simplifies its supervision, as only \textit{scene-level} label $\y$ is needed at training time, instead of the much expensive~$\ypoint$ label.

\paragraph{A pitfall of class activation mapping (CAM)-based methods -- \cref{fig:grad}} 
\ky{%
However, a noteworthy pitfall of CAM is exactly the average pooling in \cref{eq:lcam} that allowed the absence of $\ypoint$.
As shown earlier in \cref{fig:grad}, this average pooling will cause the features for points that do not belong to a certain class to remain affected by them.
This effect is further exacerbated by the fact that one typically learns a simple mapping layer, \textit{e.g.}, a linear layer in the case of \cref{eq:lcam}, which further makes it difficult to fully isolate the effect each scene-level label has on the training.
}%

\subsection{Dense Supervision from Scene-level Labels}
\label{sec:framework}
\label{sec:semantic-segmentation-module}

\ky{%
To avoid this pitfall, we propose to augment training with losses formed with careful association between labels and points.
Specifically, we form dataset-level feature clusters which we name primitives (\cref{sec:clustering}), through which we let scene-level labels to flow through~(\cref{sec:assignment}).
}%

\subsubsection{Unsupervised Feature Clustering}
\label{sec:clustering}

\ky{%
We first start by collecting point features~$\{\features_n\}$ from \textit{all} $N$ point clouds in the dataset $\dataset$ and apply K-means clustering to obtain~$\featuresprim \in \real^{K\times H}$ feature \textit{centroids} which we name \textit{primitives}\footnote{Note that we only perform this every $10$ epochs as executing K-means clustering on the whole dataset is computationally intensive.}.
While ideally with perfect features a single primitive should describe a single object class, this is something that cannot be expected during training.
Therefore, over-segment the feature space by setting $K {\gg} C$.
}%

\ky{%
Still, as we base our method on unsupervised clustering, we encourage the features to form such clusters in the feature space nearby their associated primitive.
Specifically, as in \cite{zhang2023growsp} we write
\begin{align}
  \label{eq:loss-us}
  \loss{us}(\pars) = \entropy \Big(
  \text{softmax}\big(\features\cdot({\featuresprim})^\top\big),
  \affinityprim%
  \Big)
  ,
\end{align}
where~$\affinityprim \in \{0, 1\}^{M\times K}$ is an assignment matrix induced by K-means, where each row is a one-hot vector assigning the $m$-th point to the $k$-th cluster, and $\text{softmax}(\cdot)$ normalizes rows so that they sum to one.
Optimizing this term ties points' and primitives' features so that their inner product reproduces the assignment matrix.

\paragraph{Primitive warm-up} 
Note that at the beginning of training, features are random, and K-means clusters semantically different features into the same group.
To avoid this, we refer~\cite{zhang2023growsp} to concatenate handcrafted features~\cite{rusu2008aligning}
and color to $\features$ for the first \ws{60 epochs} of training.
Details about these handcrafted features can be found in the \supplementary{}.
}%

\subsubsection{Semantic Assignment}
\label{sec:assignment}
\ky{%
We then associate each primitive with scene labels, such that the scene-level labels can be used to supervise the point features.
Specifically, we write
\begin{align}
  \label{eq:loss-match}\loss{match} & = \entropy(
  \linear(\features; \paramlinear), \: \ypoint)
  ,
\end{align}
where the per-point label $\ypoint$ is obtained for each point by figuring out which primitive each point is associated with, and if the associated primitive is again associated with a label, what that label is.
}%
To associate primitives with scene labels, one can think of various ways.
We first discuss the na\"ive linear assignment and its pitfalls, then describe our bipartite matching strategy.

\paragraph{Na\"ive assignment}
A simple way to link between clusters and scene labels could be to use a linear classifier~$\linear(\cdot; \paramlinear)$ to classify the primitives.
However, as shown in~\cref{sec:prim_match}, the linear classifier tends to predict noisy semantic assignment.

\paragraph{Bipartite assignment}
Instead, we propose to assign semantic labels by optimizing for a bipartite matching between primitives and \ky{the labels for each scene} via the Hungarian algorithm~\cite{kuhn1955hungarian}.
\ky{Specifically, to obtain $\ypoint$ we first find the association function $\pi(k)$, that maps a cluster $k$ either into one of the labels or a null class $\emptyset$ if there is no label associated to the primitive.
We thus write}
\begin{align}
  \label{eq:bipartite-matching}
  \pi(k) =
  \argmax_{ \{c:\,  c=\emptyset \, \| \, \y[c]=1\} }
  \left(
  \sum_{k=1}^{K} \mathbf{E}_{k, c}
  \right)
  ,
\end{align}
\ky{%
where the cost matrix $\mathbf{E} \in \real^{K\times C}$ is a cost matrix defining the cost of each association that we define as
}%
\begin{align}
  \mathbf{E}_{k, c}\!=\!
   \avgfeaturesprim_{k} \cdot \paramlinear_{(:, c)} , \,
  \quad
  \avgfeaturesprim_k\!=\! \frac{\sum_m \affinityprim_{m,k} \odot \features_{m,k}}{\sum_m \affinityprim_{m,k}} 
  \label{eq:cost}
  ,
\end{align} 
\ky{
where $\paramlinear_{(:, c)}$ is a linear layer. 
Note that $\avgfeaturesprim_k$ is effectively the feature representation for the $k$-th primitive that is computed for the given scene.
}
\ky{%
Then, with $\pi(\cdot)$ we now write the per-point label $\ypoint_m$ for the $m$-th point in the point cloud as
\begin{align} 
    \ypoints_m = \text{onehot}\left(\pi \left(
    \argmax_{k\in [1, K]} \left(\affinityprim_{m} \right)
    \right) \right)
    ,
\end{align}
which simply is one-hot encoding of the assignments given by $\pi(\cdot)$ and the affinity matrix $\affinityprim_{m}$ defining the point-primitive relationship.
}%

\subsubsection{Bootstrapping}
We follow the scene-level supervised methods to further boost the performance via bootstrapping~\cite{Wei_2020_CVPR,mit_2023}.
\ky{%
Typically, the bootstrapping setup involves re-training a new segmentation network with only the most confident estimates from initial training, thus aiming to reduce prediction error.
In~\citet{mit_2023}, the new network is a sparse residual U-Net~\cite{xie2020pointcontrast}, which we follow.
Existing works rely on complex confidence-based filtering steps~\cite{Wei_2020_CVPR,mit_2023} to perform bootstrapping.
In our case, we simply keep points that are estimated to adhere to the scene-level label for bootstrapping.
For the rest of the points that do not have labels, both in ours and other methods that perform bootstrapping, they are ignored and excluded from supervision.
While bootstrapping improves performance, we note that
}%
that our method outperforms all the baselines by a large margin \textit{even without bootstrapping}.

%% file: figs/framework.tex
\newcommand{\kkframework}{
  \resizebox{\linewidth}{!}{
  \centering
  \begin{overpic}[scale=1.0,percent]{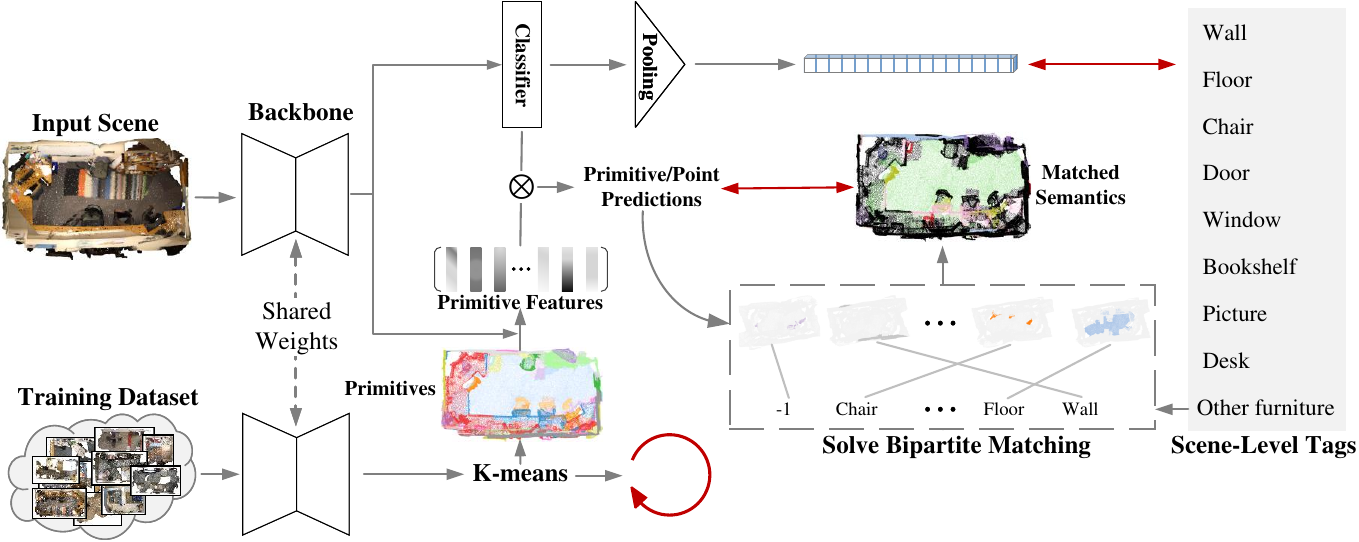}
  \put(47.3, 4.5){\textbf{\cref{eq:loss-us}}}
  \put(56.2,26.8){\textbf{\cref{eq:loss-match}}}
  \put(79.1,35.9){\textbf{\cref{eq:lcam}}}
\end{overpic}
  }
}
\begin{figure*}
  \centering
  \kkframework
  \caption{
    \textbf{Framework} -- We show the overview of our approach for weakly supervised point cloud segmentation. We leverage K-means performed on features from a point cloud backbone. We then apply bipartite matching to assign semantic labels conservatively to the found clusters, or \textit{primitives}. Finally, we propagate the assigned labels to points as a pseudo-ground truth.
  }
  \label{fig:framework}
\end{figure*}

%% file: sec/4_experiments.tex
\section{Experiments}
\label{sec:results}
Following the baselines, we benchmark our method on ScanNet~\cite{dai2017scannet} and S3DIS~\cite{armeni20163d} datasets for large-scale point cloud semantic segmentation. 
ScanNet \cite{dai2017scannet} consists of 1,201 training scenes, 312 validation scenes, and 100 test scenes, with 20 semantic classes.
S3DIS~\cite{armeni20163d} contains 272 rooms 
from 6 indoor areas and has 13 semantic categories.
We compare our approach to the existing scene-level supervised methods including PCAM~\cite{Wei_2020_CVPR}, MPRM~\cite{Wei_2020_CVPR}, MIL-Seg~\cite{ren20213d}, WYPR~\cite{ren20213d}, MIL-Trans~\cite{yang2022mil}, and the recent state-of-the-art MIT~\cite{mit_2023} and MMA~\cite{li2024multi}.
We also provide the results for fully supervised approaches including ScanNet~\cite{dai2017scannet}, PointNet++~\cite{qi2017pointnet++}, SPLATNet \cite{su2018splatnet}, KPConv~\cite{thomas2019kpconv} and MinkNet~\cite{choy20194d}.
Following the experimental setting in the baselines \cite{qi2017pointnet++, XuTowards2020, Wei_2020_CVPR, yang2022mil},  we select \texttt{Area 5} as the test scene and use other areas for training. 
We extract scene-level labels from the provided dense ground-truth labels.
We use mean Intersection over Union (mIoU) as the metric to evaluate the semantic segmentation results.

\subsection{Implementation Details}
\label{sec:impl_details}
We use MinkowskiNet~\cite{choy20194d} with voxel size of {2cm} as our point cloud backbone. The implementation of CAM baseline is provided by~\citet{ren20213d}. The number of epochs is $400$. We set $K{=}700$ and perform K-means every $10$ epochs.
As performing K-means on all the points is computationally prohibitive, we cluster points to super-voxels, following the same algorithm as in~\cite{zhang2023growsp}.
\input{tbls/scannet}
Our warm-up procedure lasts for 250 training epochs. 
We train our model on NVIDIA~3090 with batch size $6$, and use the  AdamW~\cite{loshchilov2018decoupled} optimizer with OneCycleLR~\cite{smith2018superconvergence} policy and learning rate set to $0.01$.

\subsection{Quantitative Results -- \cref{tab:scannet}}
\label{sec:quantitative}
We compare our method with scene-level supervised and fully supervised baselines.
Our approach significantly surpasses scene-level baselines -- it displays an 8.9 boost in mIoU on ScanNet and a 4.8 boost on S3DIS.
\ky{These each correspond to 29\% and 18\% relative gain, a significant performance leap.}
Furthermore, we are among the first to outperform three of the fully supervised methods, ScanNet~\cite{dai2017scannet}, PointNet++~\cite{qi2017pointnet++} and SPLATNet~\cite{su2018splatnet} on the ScanNet~\cite{dai2017scannet} dataset. 
It shows 
that weakly supervised approaches have a great potential for precise segmentation. 
We note that our method achieves significantly better results on ScanNet than~S3DIS.
\ky{%
We suspect that this is related to an issue with the dataset -- the dataset exhibits a co-occurrence bias having some of the classes in the dataset always appearing or not appearing together.
We discuss this issue in \Cref{sec:s3dis-coocurrence}.
}%

\input{figs/quali_scannet}

\subsection{Qualitative Results -- \cref{fig:scannet_quali}} 
We compare our approach qualitatively in~\cref{fig:scannet_quali} with the CAM baseline~\cite{Wei_2020_CVPR}. 
Our method performs significantly better even without retraining.
It can distinguish semantically similar objects close to each other spatially (the first row in~\cref{fig:scannet_quali}). 
Our predicted labels are also more spatially smooth than the baseline (zoom-ins), because unsupervised primitives can group close points into coherent segments. 
Furthermore, the \ws{bootstrapping} removes many erroneous supervision signals, lowering the noise level in our results in~\cref{fig:scannet_quali}.

\input{figs/showmatch}

\input{tbls/abs_label_s3dis}

\subsection{Primitive Matching -- \cref{fig:showmatch}}
\label{sec:prim_match}
We further show that our primitive matching method is critical.
We start by defining a simple baseline,~``Na\"ive matching'' which aligns primitives and semantic labels by $\linear(\cdot; \paramlinear)$. 
In~\Cref{fig:showmatch}, we visually demonstrate that our proposed bipartite matching is critical to correctly associate scene-level tags to points by comparing matched semantics with ground truth.
The classifier $\linear(\cdot; \paramlinear)$ predicts incorrect semantics for primitives, which na\"ive matching cannot rule out.
To lessen the impact of wrong predictions, we match only one primitive for each category and drop the others (black points in~\cref{fig:showmatch}).
The primitives guided by na\"ive matching may also fail in separating different semantic regions.
(\eg, table and chair in the second row), indicating that our matching mechanism also helps in unsupervised clustering.
Furthermore, some semantic tags may not correspond to any primitives in na\"ive matching due to the limitation of~$\linear(\cdot; \paramlinear)$.
For example, in the third row, na\"ive matching fails to find corresponding primitives for~``table''. 
In contrast, our method ensures that every scene-level label has a proper corresponding primitive. 

\subsection{Label Co-occurrence in S3DIS -- ~\cref{tab:abs_label_s3dis}}
\label{sec:s3dis-coocurrence}
We now show that S3DIS dataset has a label co-occurrence problem, which limits the performance of our method and other CAM-based methods in \cref{tab:scannet}.
Specifically, we observe that the~``floor'', ``ceiling'' and~``wall'' categories have the co-occurence as they exist in all scenes of S3DIS dataset (detailed in the \supplementary{}). 
When calculating semantic-related losses from~\cref{eq:lcam,eq:loss-match}, mislabeling those categories interchangeably (\eg, floor to ceiling) does not affect the loss value. 
Therefore, this semantic mapping is ill-posed, leading to an inferior and unstable training of the model. 
\ky{We observe that this is the case for both CAM and our method on this dataset.}

We demonstrate that simply removing co-occurrence with minimal manual labour can significantly improve the performance of both our method and CAM.
We randomly select only two scenes from S3DIS and remove ``floor'' in one of the scenes and ``ceiling'' in another, which guarantees that, in at least one scene, three categories don't co-occur.
We present our findings in~\cref{tab:abs_label_s3dis}, with CAM serving as a baseline.
Surprisingly, breaking the co-occurrence of labels by simply modifying \emph{two} scenes in the dataset can notably improve the performance of both methods.
\ky{%
In hindsight this is natural -- the entire premise of scene-level supervision relies that these co-occurences do not exist.
Still, as demonstrated, it is an easy thing to overlook.
We show, however, that with only a small manual effort, this is avoidable.
}%
We leave the automatic solution to the label occurrence problem as future work.

\subsection{Ablation Studies}
We perform ablation studies on ScanNet, as it is more reliable than S3DIS 
as analyzed in~\cref{sec:s3dis-coocurrence}.  
We first show that training the primitives in tandem with CAM is essential.
We then examine the importance of $\loss{us}$ and $\loss{match}$ and  search for the optimal number of primitives.

\input{tbls/abs_design}

\paragraph{Training Primitives~--~\cref{table:ab_design}}
Leveraging pre-trained primitives in the CAM-based methods is a logical idea as primitives can serve as regularizers and improve performance directly.
We implement this idea by applying the affinity matrix $\affinityprim$ to supervise the method trained on the CAM loss \cref{eq:lcam} only.
As shown in in~\cref{table:ab_design}, \ky{using pre-trained primitives only marginally}
improves the CAM performance.
\ky{Moreover,}
it is still lower than primitive clustering from~\cite{zhang2023growsp}.
\ky{%
This suggests that proper integration is necessary, and our joint training framework based on bipartite matching is essential.
}%

\input{tbls/abs_us_match}

\paragraph{Loss components -- \cref{table:abs_us_match}}
We investigate the effect~$\loss{us}$ and~$\loss{match}$ have on our method.
We show the results in~\cref{table:abs_us_match} and use~$\loss{cam}$ as a baseline. 
Adding~$\loss{us}$ to the objective leads to a mIoU increase by 5.3.
$\loss{us}$ helps to maintain primitive features and features of the corresponding points close to each other, \ie, they do not diverge throughout the training.
When using~$\loss{match}$ alone, the performance decreases by 0.7. 
However, using \textit{both}~$\loss{us}$ and~$\loss{match}$ gives the best results.
\ky{%
Having $\loss{us}$ is critical to benefiting from $\loss{match}$ and the bipartite matching, as without the centroid features diverge from the point features over training.
Note also how applying a na\"ive matching~(denotes as $\loss{match}^\star$) simply harms and does not allow this leap in performance.
}%

\input{tbls/abs_K}

\paragraph{Number of primitives -- \cref{tab:abs_k_value}}
The number of primitives~$K$ affects the performance of the proposed framework.
At low values $K{=}100$, the number of primitives is insufficient to fully capture the semantically different parts of objects in the dataset, resulting in the lowest mIoU as shown in~\cref{tab:abs_k_value}.
On the other hand, the mIoU saturates quickly as~$K$ increases. We achieve the best performance at~$K{=}700$.
When~$K$ becomes larger, the performance decreases slightly at the cost of increased training time.
More primitives result in clustering smaller regions, leading to a smaller set of points being pseudo-labelled and inferior performance.

%% file: tbls/scannet.tex
\begin{wrapfigure}{r}{0.6\linewidth}
  \centering
  \resizebox{\linewidth}{!}{
    \setlength{\tabcolsep}{8pt}
    \begin{tabular}{llcccccc}
      \toprule
       & \multirow{2}[2]{*}{Methods}        & \multicolumn{3}{c}{ScanNet~\cite{dai2017scannet}} & S3DIS~\cite{armeni20163d}                                 \\
      \cmidrule(lr){3-5} \cmidrule(l){6-6} 
       &                                                                   & Train                                             & Val                       & Test          & Test          \\

      \midrule
      \multirow{5}{*}{\rotatebox[origin=c]{90}{\small Full}}
       & ScanNet \cite{dai2017scannet}                                 & -                                                 & -                         & 30.6          & -             \\
       & PointNet++ \cite{qi2017pointnet++}                              & -                                                 & -                         & 33.9          & -             \\
       &SPLATNet \cite{su2018splatnet}                                  & -                                                 & -                         & 39.3          & -         \\
      &KPConv \cite{thomas2019kpconv}                                   & -                                                  & -                         & 68.4          & 67.1          \\
      &MinkNet \cite{choy20194d}                                         & -                                                 & 72.2                         & 73.6          & 65.4          \\
      \midrule
      \multirow{8}{*}{\rotatebox[origin=c]{90}{ \small Scene-level}}
       & PCAM \cite{Wei_2020_CVPR}$^\star$                               & 22.1                                              & -                         & -              & -             \\
       & MPRM \cite{Wei_2020_CVPR}$^\star$                               & 24.4                                              & -                         & -              & 10.3          \\
       & MIL-Seg  \cite{ren20213d}                                       & -                                                 & 20.7                      & -              & -             \\
       & WYPR   \cite{ren20213d}                                         & -                                                 & 29.6                      & 24.0           & 22.3          \\
       & MIL-Trans\cite{yang2022mil}                                     & -                                                 & 26.2                      & -              & 12.9          \\
       & MIT  \cite{mit_2023}$^\star$                                    & -                                                 & 31.6                      & 26.4           & 23.1          \\
       & MMA \cite{li2024multi}                                          & -                                                 & -                      & 30.6           &26.3          \\
       & \textbf{Ours}                                                   & \underline{39.7}                                  & \underline{38.9}          & -              & \underline{29.8}          \\
       & \textbf{Ours}$^\star$                                           & \textbf{43.2}                                                & \textbf{42.7}             & \textbf{39.5}  & \textbf{31.1} \\
      \bottomrule
    \end{tabular}
  }
  \captionof{table}{
      \textbf{Quantitative Results} -- We show the results on the ScanNet~\cite{dai2017scannet} and S3DIS~\cite{armeni20163d} datasets. Our method achieves state-of-art performance compared to other scene-level weakly supervised methods (``Scene-level'').
      We further compare with the \textit{fully supervised} methods (``Full''). 
      After the bootstrapping (denoted as $\star$), our method matches the performance of fully supervised methods.
      Note as suggested by benchmark~\cite{dai2017scannet}, we test our method with bootstrapping \textit{only once} on held-out test split. %
  }
  \label{tab:scannet}
 \vspace{-2em}
\end{wrapfigure}

%% file: figs/quali_scannet.tex
\newcommand{\qualiscannetimagewidth}{4.5cm}

\newcommand{\firstrowname}{0153_01_Toilet_0153_01}
\newcommand{\secondrowname}{0342_00_Hall_0342_00}
\newcommand{\thirdrowname}{0609_03_Lounge_0609_03}
\newcommand{\fourthrowname}{0621_00_Conference_0621_00}
\newcommand{\qualiscannetrowimage}[2]{%
    \includegraphics[width=\qualiscannetimagewidth, trim={0pt 0pt 0pt 0pt}, clip]{assets/qualitative_neurips/Fig_4_New_#1_#2_p.png}%
}

\newcommand{\versionone}{
    \begin{tikzpicture}[
        >=stealth',
        overlay/.style={
          anchor=south west, 
          draw=black,
          rectangle, 
          line width=0.8pt,
          outer sep=0,
          inner sep=0
        }
    ]
        \matrix[
        matrix of nodes, 
        column sep=0pt, 
        row sep=0pt, 
        ampersand replacement=\&, 
        inner sep=0, 
        outer sep=0
        ] (main) {
            \qualiscannetrowimage{\firstrowname}{CAM} \&
            \qualiscannetrowimage{\firstrowname}{Ours1} \&
            \qualiscannetrowimage{\firstrowname}{Ours2} \&
            \qualiscannetrowimage{\firstrowname}{GT} \\
            
            \qualiscannetrowimage{\secondrowname}{CAM} \&
            \qualiscannetrowimage{\secondrowname}{Ours1} \&
            \qualiscannetrowimage{\secondrowname}{Ours2} \&
            \qualiscannetrowimage{\secondrowname}{GT} \\
            
            \qualiscannetrowimage{\thirdrowname}{CAM} \&
            \qualiscannetrowimage{\thirdrowname}{Ours1} \&
            \qualiscannetrowimage{\thirdrowname}{Ours2} \&
            \qualiscannetrowimage{\thirdrowname}{GT} \\
            
            \qualiscannetrowimage{\fourthrowname}{CAM} \&
            \qualiscannetrowimage{\fourthrowname}{Ours1} \&
            \qualiscannetrowimage{\fourthrowname}{Ours2} \&
            \qualiscannetrowimage{\fourthrowname}{GT} \\
        };
        
        \node[below=0.1em of main-4-1.south, align=center, anchor=north]{CAM~\cite{Wei_2020_CVPR}};
        \node[below=0.1em of main-4-2.south, align=center, anchor=north]{\textbf{Ours}};
        \node[below=0.1em of main-4-3.south, align=center, anchor=north]{\textbf{Ours}$^\star$};
        \node[below=0.1em of main-4-4.south, align=center, anchor=north]{Ground Truth};
    \end{tikzpicture}
}

\begin{figure*}[t]
  \resizebox{\linewidth}{!}{\versionone}
  \caption{
    \textbf{Qualitative results} -- Our method (\textbf{Ours}), predicts more accurate and more spatially consistent labels.  In contrast, the CAM-based baseline method often produces noisy classes, which is easily visible in the first row of results. Bootstrapping~(\textbf{Ours}$^\star$) further improves the performance.
  }
  \label{fig:scannet_quali}
\end{figure*}

%% file: figs/showmatch.tex
\newcommand{\showmatchimagewidth}{4cm}

\newcommand{\showmatchrowimage}[2]{%
    \includegraphics[width=\showmatchimagewidth,trim={0pt 0pt 0pt 0pt}, clip]{assets/primitives_neurips/Fig_5_New_row#1_#2_p.png}%
}

\newcommand{\showmatchversionone}{
    \begin{tikzpicture}[
        >=stealth',
        overlay/.style={
          anchor=south west, 
          draw=black,
          rectangle, 
          line width=0.8pt,
          outer sep=0,
          inner sep=0
        }
    ]

        \matrix[
            matrix of nodes,
            column sep=0pt,
            row sep=0pt,
            ampersand replacement=\&,
            inner sep=0,
            outer sep=0,
        ] (naivematching) {
            \showmatchrowimage{1}{naive_primitives-0471_00} \&
            \showmatchrowimage{1}{naive_matched_0471_00} \\
            \showmatchrowimage{2}{naive_primitives} \&
            \showmatchrowimage{2}{naive_matched_0045_00} \\
            \showmatchrowimage{5}{naive_primitives_0013_02} \&
            \showmatchrowimage{5}{naive_matched_0013_02} \\
        };

        \matrix[
            matrix of nodes,
            column sep=0pt,
            row sep=0pt,
            ampersand replacement=\&,
            inner sep=0,
            outer sep=0,
            right=0.5em of naivematching.east
        ] (ourmatching) {

            \showmatchrowimage{1}{ours_primitives_0471_00} \&
            \showmatchrowimage{1}{ours_matched_0471_00} \\
            
            \showmatchrowimage{2}{ours_primitives_0045_00} \&
            \showmatchrowimage{2}{ours_Matched_0045_00} \\
            
            \showmatchrowimage{5}{ours_primitives_0013_02} \&
            \showmatchrowimage{5}{ours_matched_0013_02} \\
        };
        
        \matrix[
            matrix of nodes,
            column sep=0pt,
            row sep=0pt,
            ampersand replacement=\&,
            inner sep=0,
            outer sep=0,
            right=0.5em of ourmatching.east
        ] (gt) {
            \showmatchrowimage{1}{GT} \\
            \showmatchrowimage{2}{GT} \\
            \showmatchrowimage{5}{GT_0013_02} \\
        };

        \draw[gray!75,thin, dashed] ($(naivematching-1-2.north east)!0.35em!(ourmatching-1-1.north west)$) -- ($(naivematching-3-2.south east)!0.35em!(ourmatching-3-1.south west)$);

        \draw[gray!75,thin, dashed] ($(ourmatching-1-2.north east)!0.35em!(gt-1-1.north west)$) -- ($(ourmatching-3-2.south east)!0.35em!(gt-3-1.south west)$);

        \node[align=center, anchor=north] at (naivematching.south) {Na\"ive Matching};
        \node[align=center, anchor=north] at (ourmatching.south) {Our Bipartite Matching};
        \node[align=center, anchor=north] at (gt.south) {Ground Truth};
        
        \node[above=0.1em of naivematching-1-1.north, align=center, anchor=south]{Primitives};
        \node[above=0.1em of naivematching-1-2.north, align=center, anchor=south]{Matched Semantics};
        \node[above=0.1em of ourmatching-1-1.north, align=center, anchor=south]{Primitives};
        \node[above=0.1em of ourmatching-1-2.north, align=center, anchor=south]{Matched Semantics};
    \end{tikzpicture}
}

\begin{figure*}
  \centering
\resizebox{\linewidth}{!}{\showmatchversionone}
  \caption{
    \textbf{Primitive Matching} -- We show that our proposed bipartite matching~(\cref{eq:bipartite-matching}) aligns \textit{primitives} better to semantic labels than na\"ive matching.
    We color primitives randomly and matched semantics according to the class label. Black colors denote unmatched primitives or ignored points. We compare the quality to the ground truth provided in the dataset.
  }
  \label{fig:showmatch}
\end{figure*}

%% file: tbls/abs_label_s3dis.tex
\begin{wrapfigure}{r}{0.45\linewidth}
    \vspace{-4em}
    \begin{center}
        \resizebox{\linewidth}{!}{
        \begin{tabular}{lcc}
            \toprule
            Method & Initial       & No co-occurrence\\
            \midrule
            CAM    & 8.4             & 12.2 $\pm$ 0.7        \\
            Ours   & 29.8            & 33.4  $\pm$ 1.6        \\
            \bottomrule
        \end{tabular} 
        } %
    \end{center} 
    \captionof{table}{
      \textbf{Impact of label co-occurrence} -- We show that simply removing the co-occurrence of scene-level labels with a small manual labour in the S3DIS dataset significantly improves our method as well as CAM.
      \ky{We also report the mean and the standard deviation of five runs, which show that training is stable in the absence of co-occurence.}
    } 
    \label{tab:abs_label_s3dis}
    \vspace{-1.5em}
\end{wrapfigure}

%% file: tbls/abs_design.tex
\begin{wrapfigure}{r}{0.45\linewidth}
\vspace{-1.5em}
  \centering
  \resizebox{\linewidth}{!}{
   \setlength{\tabcolsep}{18pt}
    \begin{tabular}{@{}lc@{}}
      \toprule
      Primitive usage strategy                & mIoU          \\
      \midrule
      Pretrained primitives \cite{zhang2023growsp} & 25.4          \\
      CAM                                          & 19.2          \\
      CAM $+$ pretrained primitives                & 23.7          \\
      Ours                                         & \textbf{38.9} \\
      \bottomrule
    \end{tabular}
    }
    \captionof{table}{
        \textbf{Training primitives} -- We highlight the importance of joint training primitives with CAM. We compare our approach by using pre-trained primitives~\cite{zhang2023growsp}, CAM alone, and a simple combination of both.
    }
    \label{table:ab_design}
    \vspace{-2em}
\end{wrapfigure}

%% file: tbls/abs_us_match.tex
\begin{wrapfigure}{r}{0.45\linewidth}
    \vspace{-1.5em}
  \centering
  {
  \resizebox{\linewidth}{!}{
   \setlength{\tabcolsep}{18pt}
    \begin{tabular}{@{}lcc@{}}
      \toprule
      Loss                                              & mIoU          & $\Delta$ \\
      \midrule
      $\loss{cam}$                                      & 19.2          & 0.0      \\
      $\loss{cam}$+ $\loss{us}$                         & 24.5          & 5.3     \\
      $\loss{cam}$+ $\loss{match}$                      & 18.5          & $-0.7$     \\
      $\loss{cam}$+ $\loss{us}$+ $\loss{match}^{\star}$ & 21.7          & 2.5     \\
      $\loss{cam}$+ $\loss{us}$+ $\loss{match}$         & \textbf{38.9} &  \textbf{19.7}    \\
      \bottomrule
    \end{tabular}
    }
  }
  \captionof{table}{
  \textbf{Losses} -- We show how $\loss{cam}$, $\loss{us}$ and $\loss{match}$  in~\cref{eq:our-loss} affect the final performance. $\loss{match}^{\star}$ refers to the na\"ive matching loss.
  } 
  \label{table:abs_us_match}
    \vspace{-1em}
\end{wrapfigure}

%% file: tbls/abs_K.tex
\begin{wrapfigure}{r}{0.35\linewidth}
    \vspace{-1.5em}
  \centering
  \resizebox{\linewidth}{!}{
  \setlength{\tabcolsep}{6pt}
  \begin{tabular}{@{}lccccc@{}}
    \toprule
    $K$     & 100    & 400       & 700          & 1000  \\
    \midrule
    mIoU    & 28.5   & 35.6     & \textbf{38.9}      & 38.1  \\
    \bottomrule
  \end{tabular}
  }
  \caption{
  \textbf{Number of primitives} -- We compare the results between different number of primitives~$K$. We achieve the best performance with $K{=}700$.
  }
  \label{tab:abs_k_value}
    \vspace{-1em}
\end{wrapfigure}

%% file: sec/5_conclusions.tex
\section{Conclusions}
We have presented a novel method for weakly supervised point cloud semantic segmentation which uses unsupervised primitives to conservatively propagate scene-level annotations to points, via bipartite matching.
Our approach achieves new state-of-the-art weakly-supervised point cloud segmentation and performs competitively with fully supervised methods while being much more label efficient.
Additionally, we have demonstrated the label co-occurrence problem in the S3DIS dataset~\cite{armeni20163d}, which impacts this research field extensively, and can be solved easily with small manual labour.
We hope our work will inspire future weakly supervised point cloud segmentation research.

%% file: sec/6_acks.tex
\section*{Acknowledgements}
This work was supported in part by the National Natural Science Foundation of China under Grant 42201481, the Scientific Research Foundation of Hunan Education Department under Grant 21B0332, in part by the Science and Technology Plan Project Fund of Hunan Province under Grant 2023JJ40024. We gratefully acknowledge the support of NVIDIA Corporation with the donation of the A100 GPU used for this research.  Lastly, we would like to thank Yuhe Jin for the insightful discussions. 

%% file: sec/X_suppl.tex
\section{Appendix}
\setcounter{page}{1}

We provide additional implementation details in~\cref{sec:supp_impl_details}, more qualitative and quantitative results in~\cref{sec:more_res}, a description of the label co-occurrence issue in~\cref{sec:s3dis_stats}, and analysis of CAM's limitations in~\cref{sec:problem_def_cam}. 
\subsection{Additional implementation details}
\label{sec:supp_impl_details}
\subsubsection{Supervoxel construction in K-means clustering}
We construct supervoxels to accelerate K-means clustering as presented in~\citet{zhang2023growsp}.    
Specifically, we combine VCCS~\cite{papon2013voxel} and region growing~\cite{adams1994seeded} to construct supervoxels for all the scenes. If a growing region covers a VCCS patch, we merge the VCCS patch into the former. We implement these methods based on Point Cloud Library~\cite{Rusu_ICRA2011_PCL} and use the same parameter settings as in~\citet{zhang2023growsp}. 
Note that the supervoxel construction is necessary only during training. 

\subsubsection{Handcrafted features during warm-up}
In~\cref{sec:clustering}, we mention regularizing the primitive clustering with handcrafted features at the beginning of the training. 
Specifically, we follow~\citet{zhang2023growsp} to compute the average RGB and PFH~\cite{rusu2008aligning} features for each supervoxel, and then concatenate them with features from the backbone, forming the input feature of K-means.

\subsubsection{No matching loss during warm-up stage}
Note we drop the $\loss{match}$ from~\cref{eq:loss-match} during the warm-up stage -- first 60 training epochs. 
This is because our matching objective relies on the quality of the trained classifier $\linear$ whereas, during the warm-up stage, the prediction of $\linear$ is noisy. 
We observe that, without doing so, the model will be stuck with the incorrect assignment, harming the final performance.

\input{tbls/scannet_perclass}
\input{tbls/s3dis_perclass}

\subsection{More results}
\label{sec:more_res}
\subsubsection{Per-class results~--~\cref{tab:res_cls_seg} and~\cref{tab:s3dis_seg}}
We present in~\cref{tab:res_cls_seg} and~\cref{tab:s3dis_seg} extended versions of our quantitative results and show mIoU score for each semantic class existing in the ScanNet~\cite{dai2017scannet} and S3DIS~\cite{armeni20163d} datasets. 
In ScanNet, our method excels in most of the categories, notably in ``sofa'', ``table'', ``chair'' and ``floor''. However, it easily confuses classes related to the ``wall'' such as ``window'' or ``picture'' in ScanNet. Similarily, in S3DIS, it fails to segment classes such as ``beam'' and ``board''. 
We suspect that points of those classes are rare in data, which potentially degrades our unsupervised primitives and also other methods.  
Nevertheless, our method still performs best among other baselines on average.

\input{tbls/analysis_label_diversity_s3dis}

\subsection{Label Co-occurrence in S3DIS -- \cref{Tab:label_div_s3dis}}
\label{sec:s3dis_stats}
To show label co-occurrence in S3DIS dataset, we statistic the frequency of scene-wise semantic labels appearing on the S3DIS training dataset in~\cref{Tab:label_div_s3dis}. The ``ceiling'',``floor'' and ``wall'' appear simultaneously in all the scenes, which is impossible to $\loss{cam}$ to work. 
In~\cref{sec:s3dis-coocurrence}, we address this issue by randomly removing ``ceiling'' and ``floor'' in different scenes to break the co-ocurrence. 

\subsection{Analysis of CAM's limitation}
\label{sec:problem_def_cam}
We ground the limitations existing in CAM-based methods~\cite{Wei_2020_CVPR} in a more formal framework. We present the visualization of that framework in~\cref{fig:grad}.

The CAM-based loss, $\loss{cam}$ in~\cref{eq:lcam}, has the one-to-many mapping between points and semantic labels---the model is trained to assign multiple labels to a single point.
As a result, it incorrectly maximizes the class score of points to the non-belonging classes.

We present the gradient issue of CAM by calculating the gradient sign of point-wise class scores to different classes where the positive gradient encourages points to be categorized as the class and vice-versa.
Specifically, we calculate the point sign of point scores derived from \textit{dense supervision} \Eq{ldense} as
\begin{align}
  \text{sign}\left(
  \frac{\partial \scores_{m,c}}{\partial{\ypoints_{m,c}}}
  \right) =
  \begin{cases}
    +, & \text{if }\ypoints_{m,c}=1 \\
    -, & \text{otherwise}
  \end{cases}
  \label{eq:sign_dense}
\end{align}
where $\ypoints_{m,c}$ is the $m$-th point's label at $C$-th class. Due to the one-hot label $\ypoint$, only one class contributes the positive gradient, contracting $m$-th point closer to one of the classes.

Similarly, we also calculate the gradient sign of \textit{scene-level supervision} using CAM derived from \Eq{lcam} and \Eq{mil} as
  \begin{align}
    \text{sign}\left(
    \frac{\partial \scores_{m,c}}{\partial \y_c}
    \right) =
    \begin{cases}
      +, & \text{if }\y_c=1, \\
      -, & \text{otherwise}
    \end{cases}
    \label{eq:sign_cam}
  \end{align}
  where $\y_c$ is the multi-hot global label shared between  $M$ points.
  Thus, in contrast to the dense case in \Eq{sign_dense}, CAM encourages a single point to be classified into multiple classes---despite a single point belonging to a single class, limiting the performance of CAM.

\subsection{Broader impact}
\label{sec:broad_impact}
Our approach is an important milestone towards democratizing low-cost, high-quality point cloud segmentation. We expect that in the near future, we will see the rise of weakly supervised methods that, by means of large language models~\cite{xu2023pointllm}, can achieve performance greater than that of fully supervised approaches.

%% file: tbls/scannet_perclass.tex
 \begin{table}[h]
    \caption{{\bf Per-class semantic segmentation results on ScanNet} -- Our method outperforms the existing state-of-the-art methods~\cite{Wei_2020_CVPR,ren20213d,yang2022mil,mit_2023} by a significant margin. ``sc'' refers to the ``shower curtain''. ``$\star$'' indicates the bootstrapping strategy. %
    }
    \centering
    \resizebox{\linewidth}{!}
    {
    \begin{tabular}{lcccccccccccccccccccccc}
   \toprule
    Method & eval. & wall & floor & cabinet & bed & chair & sofa & table & door & window & shelf & picture & counter & desk & curtain & fridge & sc & toilet & sink & bathtub & other &IoU \\
    \midrule
    PCAM$^\star$~\cite{Wei_2020_CVPR} & train & 54.9 & 48.3 & 14.1 & 34.7 & 32.9 & 45.3 & 26.1 & 0.6 & 3.3 & 46.5 & 0.6 & 6.0 & 7.4 & 26.9 & 0.0 & 6.1 & 22.3 & 8.2 & 52.0 & 6.1 & 22.1 \\
    MPRM$^\star$~\cite{Wei_2020_CVPR} & train & 47.3 & 41.1 & 10.4 & 43.2 & 25.2 & 43.1 & 21.5 & 9.8 & 12.3 & 45.0 & 9.0 & 13.9 & 21.1 & 40.9 & 1.8 & 29.4 & 14.3 & 9.2 & 39.9 & 10.0 & 24.4 \\
    \midrule
    MIL-seg~\cite{ren20213d}& val & 36.4  & 36.1  &  13.5  & 37.9  & 25.1  & 31.4  & 9.6  &  18.3  & 19.8 & 33.1  &  {\bf 7.9}  & 20.3 & 21.7 & 32.5 & 6.4 & 14.0 & 7.9 & 14.7 & 19.4 & 8.5 & 20.7 \\
    WYPR~\cite{ren20213d} & val &  58.1 & 33.9 & 5.6 &  56.6 & 29.1 &   45.5 & 19.3 & 15.2 &  {\bf 34.2} &  33.7 & 6.8 &  {\bf 33.3} & 22.1 & {\bf 65.6} &  {\bf6.6} &  36.3 & 18.6 &  24.5 &  39.8 & 6.6 & 29.6 \\
    MIL-Trans~\cite{yang2022mil}  & val &  - & - & - & - & - &  - & - & - &  - &  - &- & - &- & - & - &  - & -  &-  & - &  - & 26.2\\
    MIT$^\star$~\cite{mit_2023}     & val & -  &- & -  & - & - &  - & - & - &  - &  - &- & - &- & - & - &  - &-   & - & - &  - & 31.1\\
    Ours & val &  63.3  & {\bf87.9}  & 26.0 &  39.2 & 60.0 & 52.5 & 48.2 & 18.8 & 32.1 & 51.7 & 0.3  &  32.3 &  21.0 & 41.0 & 1.1 &  51.9 &  61.1 & 19.3 & 53.3 &  17.0 & 38.9 \\
    Ours$^\star$ & val & {\bf 66.5} &  87.7 &  {\bf28.8} &   46.7 & {\bf 63.0} &  {\bf 57.6} & {\bf 50.9} & {\bf21.4} & 34.0 &  {\bf 54.8} & 0.0 & {\bf34.7}  & 26.6 & 47.6 & 0.8 &  {\bf 55.5} & {\bf 74.8} & 19.5 &  {\bf 61.1} & {\bf 22.2} & {\bf 42.7} \\

    \bottomrule
    \end{tabular}
    }
    \vspace{-2em}
    \label{tab:res_cls_seg}
\end{table}

%% file: tbls/s3dis_perclass.tex
 \begin{table}[h]
 	\caption{{\bf Per-class semantic segmentation results on S3DIS} --  Our method performs best in terms of the average mIoU. ``$\star$'' indicates the bootstrapping strategy. Please note that the existing state-of-the-art methods~\cite{Wei_2020_CVPR,ren20213d,yang2022mil,mit_2023} do not report per-class evaluations.}
  
 	\centering

  \resizebox{\linewidth}{!}{
     	\begin{tabular}{lcccccccccccccc}
 	    \toprule
 	Method                       & ceiling   & floor   & wall   & beam   &column   & window  & door  & table  & chair  &sofa   & bookcase  & board  &clutter   &mIoU \\

    \midrule
MPRM$^\star$~\cite{Wei_2020_CVPR} & -         &-        & -      & -       & -      & -       & -     & -      & -      & -     & -         & -      & -        & 10.3 \\
WYPR~\cite{ren20213d}             & -         &-        & -      & -       & -      & -       & -     & -      & -      & -     & -         & -      & -        & 22.3 \\
MIL-Trans~\cite{yang2022mil}      & -         &-        & -      & -       & -      & -       & -     & -      & -      & -     & -         & -      & -        & 12.9 \\
MIT$^\star$ ~\cite{mit_2023}      & -         &-        & -      & -       & -      & -       & -     & -      & -      & -     & -         & -      & -        & 23.1 \\
MMA~\cite{li2024multi}  & -         &-        & -      & -       & -      & -       & -     & -      & -      & -     & -         & -      & -        & 26.3 \\
Ours                  & 79.4      & 89.6    & 55.4   & 0.0     &\bf6.3  & \bf5.2     &\bf 12.8 & 51.3  & \bf13.4   & 15.9   & 37.5      & 0.0    & 20.7      & 29.8 \\
Ours$^\star$     &\bf 83.0   &\bf91.2     &\bf59.5   &\bf 0.0  & 5.3     & 4.8    & 10.7     & 55.1   & 10.3   & \bf24.0   & \bf38.5      & 0.0    & \bf21.2      & \bf31.1 \\
 	     \bottomrule
 	\end{tabular}
  }
  \vspace{-1em}
 	\label{tab:s3dis_seg} 
 \end{table}

%% file: tbls/analysis_label_diversity_s3dis.tex
\begin{table}[h]
    \setlength{\tabcolsep}{4pt}
    \caption{
        {\bf Label co-occurrence matrix} -- We show that ``ceiling'',``floor'' and ``wall'' appear in all scenes in the S3DIS~\cite{armeni20163d} training set. Using those labels hinders the performance of our method.
    }
    \centering
    \resizebox{\linewidth}{!}
    {
    \begin{tabular}{lccccccccccccc}
    \toprule
      & ceiling & floor & wall & beam   & column & window & door & table & chair & sofa & bookcase & board & clutter \\
      \midrule
    ceiling & \textbf{204} & \textbf{204} & \textbf{204} & 104 & 86 & 80 & 190 & 138 & 141 & 22 & 126 & 77 & 202  \\ 
    floor & \textbf{204} & \textbf{204} & \textbf{204} & 104 & 86 & 80 & 190 & 138 & 141 & 22 & 126 & 77 & 202  \\ 
    wall & \textbf{204} & \textbf{204} & \textbf{204} & 104 & 86 & 80 & 190 & 138 & 141 & 22 & 126 & 77 & 202  \\ 
    beam & 104 & 104 & 104 & 104 & 60 & 65 & 97 & 91 & 87 & 11 & 81 & 59 & 104  \\ 
    column & 86 & 86 & 86 & 60 & 86 & 63 & 83 & 74 & 77 & 13 & 75 & 49 & 85  \\ 
    window & 80 & 80 & 80 & 65 & 63 & 80 & 79 & 78 & 79 & 13 & 76 & 56 & 80  \\
    door & 190 & 190 & 190 & 97 & 83 & 79 & 190 & 133 & 136 & 20 & 124 & 76 & 189  \\
    table & 138 & 138 & 138 & 91 & 74 & 78 & 133 & 138 & 129 & 21 & 113 & 73 & 138  \\ 
    chair & 141 & 141 & 141 & 87 & 77 & 79 & 136 & 129 & 141 & 21 & 112 & 74 & 140  \\ 
    sofa & 22 & 22 & 22 & 11 & 13 & 13 & 20 & 21 & 21 & 22 & 15 & 9 & 22  \\
    bookcase & 126 & 126 & 126 & 81 & 75 & 76 & 124 & 113 & 112 & 15 & 126 & 67 & 126  \\ 
    board & 77 & 77 & 77 & 59 & 49 & 56 & 76 & 73 & 74 & 9 & 67 & 77 & 77  \\ 
    clutter & 202 & 202 & 202 & 104 & 85 & 80 & 189 & 138 & 140 & 22 & 126 & 77 & 202  \\ 
    \bottomrule
    \end{tabular}}
    \label{Tab:label_div_s3dis}
\end{table}